\pdfoutput=1
\documentclass[11pt]{article}

\usepackage[final]{acl}

\usepackage{times}
\usepackage{latexsym}
\usepackage{amsmath}
\usepackage{booktabs} 
\usepackage[T1]{fontenc}
\usepackage[utf8]{inputenc}

\usepackage{microtype}

\usepackage{inconsolata}

\usepackage{graphicx}

\usepackage{multirow}
\usepackage{booktabs}
\usepackage{colortbl}
\usepackage{array}

\usepackage{booktabs} % 负责处理表格线条
\usepackage{multirow} % 负责表格的多行合并

\usepackage{subcaption}
\usepackage[ruled,vlined]{algorithm2e}

\usepackage{amsmath}
\usepackage{array}
\usepackage{booktabs}
\definecolor{lightgray}{gray}{0.9}
\definecolor{darkgray}{gray}{0.7}
\usepackage{xcolor}
\definecolor{burntorange}{RGB}{204, 85, 0} % Example RGB values

\usepackage{booktabs}
\usepackage{multirow}
\usepackage{caption}
\definecolor{darkgreen}{rgb}{0.0, 0.5, 0.0}
\usepackage{algpseudocode}
\usepackage{algpseudocode}
\usepackage{adjustbox}

\usepackage{times}
\usepackage{bm}

\usepackage{amsmath} 
\usepackage{latexsym}
\usepackage{amsfonts}
\usepackage{amssymb}

\usepackage{xcolor}
\usepackage{tabularx}

\usepackage{xspace}
\usepackage{booktabs}
\usepackage{multirow}
\usepackage{graphicx}
\usepackage{booktabs}
\usepackage{colortbl}
\usepackage{float}  % Add this line to use the H float option
\usepackage{xcolor}
\usepackage{booktabs}
\usepackage{multirow}
\usepackage{colortbl}
\usepackage{amsmath} % for math commands like \sum
 \usepackage{cleveref}
\usepackage{multirow}
\usepackage{siunitx}
\usepackage{adjustbox}
\usepackage{xcolor}
\usepackage{multirow} 
\usepackage{colortbl}
\usepackage{booktabs}
\usepackage[T1]{fontenc}
\usepackage[utf8]{inputenc}
\usepackage{microtype}
\usepackage{inconsolata}
\usepackage{tikz}
\usepackage[utf8]{inputenc}  % 添加对 UTF-8 编码的支持
\usepackage{subcaption} 
\usepackage{breqn}
\usepackage{array}
\usepackage{graphicx}
\usepackage{booktabs}
\usepackage{caption}
\usepackage{multirow}
\newcolumntype{M}[1]{>{\centering\arraybackslash}m{#1}}

\usepackage{amsmath}
\usepackage{algpseudocode}
\usepackage{dsfont}
\usepackage{amsmath}
\usepackage{amsfonts}
\usepackage{etoc}
\usepackage{tcolorbox}
\usepackage{amsmath} % 提供高级数学环境
\usepackage{amssymb} % 提供额外的数学符号（如 \mathbb）
\usepackage{dblfloatfix}

\usepackage{booktabs} 
\usepackage{amsmath} % 用于数学符号

\usepackage{booktabs}

\usepackage{xcolor}
\usepackage{tcolorbox}
\usepackage{todonotes}

\usepackage{graphicx}
\usepackage{natbib}
\usepackage{hyperref}

\title{LoRA-MGPO: Mitigating Double Descent in Low-Rank Adaptation via Momentum-Guided Perturbation Optimization}

\author{
        Yupeng Chang$^{1}$ \quad Chenlu Guo$^{1}$ \quad  Yi Chang$^{1,2,3}$ \quad Yuan Wu$^{1}$\footnotemark[1] \\
        $^{1}$School of Artificial Intelligence, Jilin University \\
        $^{2}$Engineering Research Center of Knowledge-Driven Human-Machine Intelligence, MOE, China \\
        $^{3}$International Center of Future Science, Jilin University\\
       \{changyp23, guocl23\}@mails.jlu.edu.cn, \{yichang, yuanwu\}@jlu.edu.cn \\   
}

\begin{document}
\etocdepthtag.toc{chapter}
\etocsettagdepth{chapter}{none}
\etocsettagdepth{appendix}{none}

\maketitle
\renewcommand{\thefootnote}{\fnsymbol{footnote}}
\footnotetext[1]{Corresponding authors}

\begin{abstract}
Parameter-efficient fine-tuning (PEFT), particularly Low-Rank Adaptation (LoRA), adapts large language models (LLMs) by training only a small fraction of parameters. However, as the rank of the low-rank matrices used for adaptation increases, LoRA often exhibits an unstable "double descent" phenomenon, characterized by transient divergence in the training loss, which delays convergence and impairs generalization by causing instability due to the attraction to sharp local minima. To address this, we introduce \textbf{LoRA-MGPO}, a framework that incorporates Momentum-Guided Perturbation Optimization (MGPO). MGPO stabilizes training dynamics by mitigating the double descent phenomenon and guiding weight perturbations using momentum vectors from the optimizer's state, thus avoiding dual gradient computations. Additionally, an adaptive normalization scheme scales the magnitude of perturbations based on an exponential moving average (EMA) of gradient norms, further enhancing stability. While EMA controls the magnitude of the perturbations, MGPO guides their direction, ensuring a more stable optimization trajectory. Experiments on a suite of natural language understanding and generation benchmarks show that LoRA-MGPO consistently achieves superior performance over LoRA and other PEFT methods. The analysis indicates that LoRA-MGPO leads to smoother loss curves, faster convergence, and improved generalization by stabilizing the training process and mitigating the attraction to sharp minima. 
% The code is publicly available at \url{https://github.com/llm172/LoRA-MGPO}.
\end{abstract}

\begin{figure*}
    \centering
    \includegraphics[width = 0.98\linewidth]{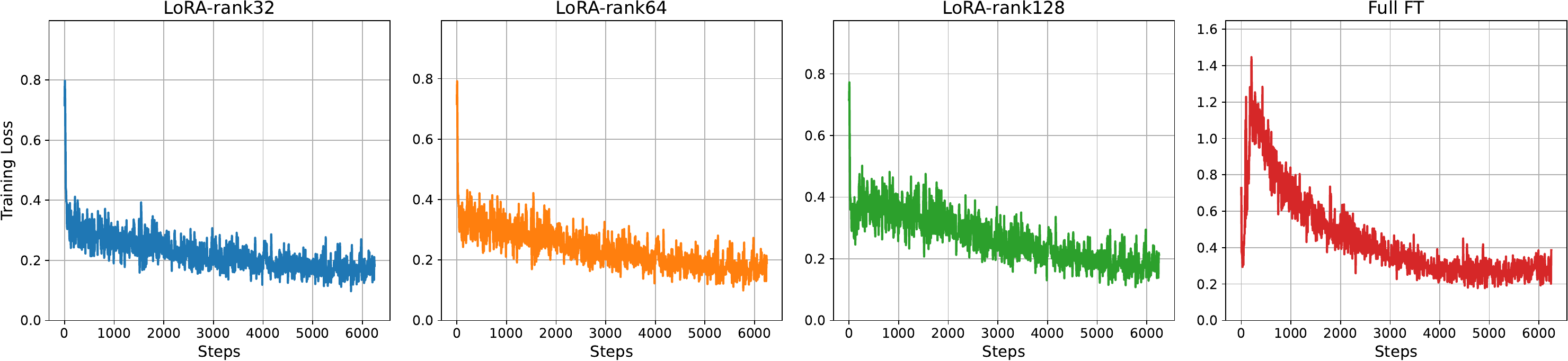}
    \caption{Training loss curves of Full FT and LoRA \cite{hu2021lora} methods with LLaMA-2-7B \citep{touvron2023llama} on the MetaMathQA dataset \citep{yu2024metamath}. For LoRA, rank ($r$) and alpha ($\alpha$) are set to the same values ($r = \alpha \in \{32, 64, 128\}$), with a fixed learning rate of $5e-4$.}
    \label{fig:fig_head}
\end{figure*}

%%%%%%%%%%%%%%%%%%%%%%%%%%%%%%%%%%%%%%%%%%

\section{Introduction}
\label{sec: intro}

Large language models (LLMs) have driven significant advancements in natural language processing, establishing new performance benchmarks on tasks ranging from text generation to semantic understanding \citep{chang2024survey,wei2022emergent}. However, the conventional method of full-parameter fine-tuning (Full FT) requires updating billions of parameters, incurring prohibitive memory and computational costs. To overcome this limitation, parameter-efficient fine-tuning (PEFT) methods have emerged as an effective alternative, enabling efficient adaptation by optimizing only a small subset of model parameters \citep{lester2021power,fu2023effectiveness}.

Among these methods, Low-Rank Adaptation (LoRA) \citep{hu2021lora} is distinguished by its computational efficiency and architectural simplicity. LoRA approximates the weight update matrix $\Delta W$ as a low-rank decomposition, where the original pre-trained weights $W_0$ remain frozen. The trainable matrices $B$ and $A$, with rank $r \ll \min(m,n)$, drastically reduce the number of trainable parameters, improving efficiency without altering the model architecture.

Despite its efficiency, LoRA's training dynamics can be unstable. As shown in Figure~\ref{fig:fig_head}, fine-tuning LLaMA-2-7B \citep{touvron2023llama} on MetaMathQA \citep{yu2024metamath} often exhibits a "double descent" trajectory with initial convergence, transient divergence, and eventual stabilization. This phenomenon worsens with higher ranks and is not unique to LoRA; Full FT can exhibit even more severe double descent, highlighting the general challenge of stabilizing fine-tuning in high-capacity models \citep{nakkiran2019deepdoubledescentbigger}. Such non-monotonic behavior delays convergence and impairs generalization due to unstable gradients and the attraction to sharp local minima \citep{li2024flat}.

Addressing these stability issues is crucial. Sharpness-Aware Minimization (SAM) \citep{foret2020sharpness} improves generalization by seeking flatter minima. However, its application is hindered by the dual gradient computation requirement, which doubles the training cost \citep{becker2024momentum, li2024revisiting}. More efficient variants like momentum-guided SAM reuse optimizer states to avoid this overhead but may not guarantee stable convergence. To further enhance stability, complementary techniques such as applying an exponential moving average (EMA) to smooth optimization dynamics have been shown to suppress parameter oscillations and improve convergence in certain scenarios \citep{wang2021rethinking}.

Building on these insights, we propose \textbf{LoRA-MGPO}, a novel framework that integrates Momentum-Guided Perturbation Optimization (MGPO) into LoRA to mitigate the detrimental effects of double descent. Our contributions are twofold:
\begin{enumerate}
    \item \textbf{Mitigating Double Descent:} MGPO stabilizes training by addressing double descent, typically observed at higher ranks in LoRA. By reusing momentum vectors, it guides weight perturbations towards flatter minima, preventing transient divergences in loss.
    \item \textbf{Adaptive Perturbation Normalization:} MGPO introduces an adaptive scheme that scales perturbation magnitude based on an exponential moving average (EMA) of gradient norms, decoupling perturbation intensity from optimization dynamics and further enhancing stability.
\end{enumerate}
We evaluate LoRA-MGPO on a suite of natural language understanding (NLU) and generation (NLG) benchmarks. Our results show that it consistently achieves superior performance over standard LoRA and other state-of-the-art PEFT methods. Crucially, we demonstrate that LoRA-MGPO effectively mitigates the double descent phenomenon, leading to more stable training dynamics, smoother loss curves, and faster convergence, all of which contribute to better generalization and the avoidance of sharp minima.

%%%%%%%%%%%%%%%%%%%%%%%%%%%%%%%%%%%%%%

%%%%%%%%%%%%%%%%%%%%%%%%%%%%%%%%%%%%%%%%%%%%%%%%%

\section{Method}
\label{sec:method}

In this section, we first provide a concise overview of the Low-Rank Adaptation (LoRA) framework. We then introduce \textbf{LoRA-MGPO}, an extension of LoRA that integrates Momentum-Guided Perturbation Optimization (MGPO) to enhance its stability and efficiency. We describe how MGPO reuses optimizer momentum for guided perturbations of the trainable parameters and incorporates an adaptive normalization scheme to stabilize training.

\subsection{Review of LoRA}
\label{sec:review_lora}

While full fine-tuning directly updates the entire pre-trained weight matrix $W_0 \in \mathbb{R}^{m \times n}$, its prohibitive computational cost makes it impractical for large-scale models. Low-Rank Adaptation (LoRA) \citep{hu2021lora} offers a parameter-efficient alternative. LoRA freezes $W_0$ and injects a trainable low-rank decomposition, $\Delta W = BA$, where $B \in \mathbb{R}^{m \times r}$ and $A \in \mathbb{R}^{r \times n}$ are trainable matrices with rank $r \ll \min(m, n)$. The weight update is incorporated into the forward pass as:
\begin{equation}
    Y = X(W_0 + \frac{\alpha}{r} BA),
\end{equation}
where $X$ is the input, $\alpha$ is a scaling hyperparameter, and $r$ is the rank of the decomposition. Typically, $A$ is initialized with a Kaiming normal distribution, and $B$ with zeros. While effective, LoRA can suffer from training instability, particularly the double descent phenomenon, when $r$ increases without appropriate optimization strategies to maintain stability \citep{li2024flat}.

\subsection{LoRA with Momentum-Guided Perturbation Optimization}
\label{sec:lora_mgpo}

To address the training instabilities in LoRA, we propose \textbf{LoRA-MGPO}, which integrates Momentum-Guided Perturbation Optimization (MGPO). Inspired by Sharpness-Aware Minimization (SAM), MGPO is redesigned for computational efficiency and parameter efficiency. It directly perturbs the trainable LoRA parameters by reusing the optimizer's first-moment estimate, guiding the perturbations toward stable directions. Additionally, MGPO incorporates adaptive normalization to dynamically scale the perturbation, enhancing training stability.

\subsubsection{Motivation: SAM for LoRA and Its Limitations}
The goal of SAM \citep{foret2020sharpness} is to find parameters in flat loss regions to improve generalization. A direct application to LoRA would involve perturbing the full weight matrix, solving $\min_{A, B} \max_{\|\epsilon\|_F \leq \rho} \mathcal{L} \left( W_0 + BA + \epsilon \right)$. This approach is ill-suited for PEFT due to two critical flaws: (1) its dual gradient computation requirement doubles the training cost, and (2) creating and storing the full-space perturbation $\epsilon$ counteracts the memory savings of LoRA. MGPO is explicitly designed to resolve these inefficiencies.

\subsubsection{Momentum-Guided Perturbation of LoRA Parameters}
MGPO achieves the stability benefits of SAM by perturbing the trainable parameters $\theta = (A, B)$ directly, using information readily available in the optimizer's state. At each training step $t$, instead of computing a new gradient for the perturbation direction, it reuses the optimizer's first-moment vector (momentum) from the previous step, $\boldsymbol{m}_{t-1}$. The optimization objective is:
\begin{equation}
\min_{\theta} \mathcal{L}(\theta_t + \epsilon_{\theta_t}),
\label{eq:mgpo_opt}
\end{equation}
where the perturbation $\epsilon_{\theta_t}$ applied to the LoRA parameters $\theta_t = (A_t, B_t)$ is constructed using the state from step $t-1$:
\begin{equation}
\epsilon_{\theta_t} = \rho \cdot \frac{\boldsymbol{m}_{t-1}}{\|\boldsymbol{m}_{t-1}\|_2} \cdot \frac{1}{\bar{g}^{(t-1)}}.
\label{eq:mgpo_perturbation}
\end{equation}
Here, $\rho$ is the perturbation radius. Using the historical momentum vector is a deliberate design choice, as it represents a smoothed average of past gradients, filtering out the noise from any single mini-batch and providing a more stable direction for assessing landscape sharpness. This vector is maintained by the optimizer itself. After computing the gradient on the perturbed parameters, the momentum for the current step is updated as:
\begin{equation}
\boldsymbol{m}_t = \mu \boldsymbol{m}_{t-1} + \nabla_{\tilde{\theta}_t} \mathcal{L}.
\label{eq:momentum_update}
\end{equation}
The decay factor $\mu$ (e.g., `beta1` in AdamW) is reused from the optimizer’s standard settings. The scalar $\bar{g}^{(t-1)}$ is a global normalization factor, detailed next. This formulation entirely avoids the second gradient computation and any operations in the full weight space.

\paragraph{Two-Stage Update Mechanism}
MGPO is implemented efficiently within each training step $t$. First, using the state from step $t-1$, we compute the perturbation $\epsilon_{\theta_t}$ and apply it to the current parameters $\theta_t$ to get a perturbed version, $\tilde{\theta}_t$:
\begin{equation}
\tilde{\theta}_t = \theta_t + \epsilon_{\theta_t}.
\label{eq:perturbed_init}
\end{equation}
Second, the loss and its gradient are computed with respect to these perturbed parameters: $\nabla_{\tilde{\theta}_t} \mathcal{L}$. This single gradient is then used by the optimizer to update both the original parameters from $\theta_t$ to $\theta_{t+1}$ and the momentum from $\boldsymbol{m}_{t-1}$ to $\boldsymbol{m}_{t}$. For inference, the final, unperturbed parameters $\theta_T$ are used.

\subsubsection{Adaptive Perturbation Normalization}
\label{sec:adaptive_norm}
To ensure robustness across training stages, we introduce an Adaptive Perturbation Normalization (APN) scheme. The normalization factor $\bar{g}^{(t)}$ used in Equation~\ref{eq:mgpo_perturbation} is a scalar computed via an exponential moving average (EMA) of the global L2-norm of the LoRA parameter gradients. Following the principle of using the actually computed gradient, the update rule is:
\begin{equation}
\bar{g}^{(t)} = \beta \bar{g}^{(t-1)} + (1-\beta)\|\nabla_{\tilde{\theta}_t}\mathcal{L}\|_2,
\label{eq:ema}
\end{equation}
where $\beta$ is the EMA decay rate. This mechanism makes the perturbation scale-invariant relative to the gradient dynamics. For instance, during early training with large gradients, the normalization factor increases, reducing the effective perturbation size to prevent destabilization. Conversely, in later stages, it ensures the perturbation remains sufficiently large to be effective. This adaptive scaling enhances training stability.

%%%%%%%%%%%%%%%%%%%%%%%%%%%%%%%%%%%%%%

\begin{table*}[htb!]
  \centering
  \caption{Performance of T5-Base on five GLUE tasks, comparing LoRA-MGPO with full fine-tuning and other LoRA variants (rank $r=8$). Scores are reported for the primary metric of each task, averaged over 3 runs, with standard deviations shown in subscripts. \textbf{Bold} indicates the best score, while \underline{underlining} denotes the second best.}

  \begin{tabular}{l|ccccc|c}
    \toprule
    \textbf{Method} & \textbf{MNLI} & \textbf{SST2} & \textbf{CoLA} & \textbf{QNLI} & \textbf{MRPC} & \textbf{Avg} \\
    \midrule
    \rowcolor{gray!10} \textit{Train Size} & \textit{393k} & \textit{67k} & \textit{8.5k} & \textit{105k} & \textit{3.7k} & \\
    \midrule
    Full FT & \underline{86.33$_{\pm0.00}$} & \textbf{94.75$_{\pm0.21}$} & 80.70$_{\pm0.24}$ & 93.19$_{\pm0.22}$ & 84.56$_{\pm0.73}$ & 87.91 \\
    LoRA  & 85.30$_{\pm0.04}$ & 94.04$_{\pm0.11}$ & 69.35$_{\pm0.05}$ & 92.96$_{\pm0.09}$ & 68.38$_{\pm0.01}$ & 82.08 \\
    \hline
    \rowcolor{gray!20}\multicolumn{7}{c}{\textit{LoRA Variants with Modified Structure}}    \\
    DoRA  & 85.67$_{\pm0.09}$ & 94.04$_{\pm0.53}$ & 72.04$_{\pm0.94}$ & 93.04$_{\pm0.06}$ & 68.08$_{\pm0.51}$ & 82.57 \\
    AdaLoRA & 85.45$_{\pm0.11}$ & 93.69$_{\pm0.20}$ & 69.16$_{\pm0.24}$ & 91.66$_{\pm0.05}$ & 68.14$_{\pm0.28}$ & 81.62 \\
    \midrule
    \rowcolor{gray!20}\multicolumn{7}{c}{\textit{LoRA Variants with Original Structure}}    \\
    PiSSA & 85.75$_{\pm0.07}$ & 94.07$_{\pm0.06}$ & 74.27$_{\pm0.39}$ & 93.15$_{\pm0.14}$ & 76.31$_{\pm0.51}$ & 84.71 \\
    rsLoRA & 85.73$_{\pm0.10}$ & 94.19$_{\pm0.23}$ & 72.32$_{\pm1.12}$ & 93.12$_{\pm0.09}$ & 52.86$_{\pm2.27}$ & 79.64 \\
    LoRA+ & 85.81$_{\pm0.09}$ & 93.85$_{\pm0.24}$ & 77.53$_{\pm0.20}$ & 93.14$_{\pm0.03}$ & 74.43$_{\pm1.39}$ & 84.95 \\
    LoRA-GA & 85.70$_{\pm0.09}$ & 94.11$_{\pm0.18}$ & 80.57$_{\pm0.20}$ & 93.18$_{\pm0.06}$ & 85.29$_{\pm0.24}$ & 87.77 \\
    \midrule
   \rowcolor{green!20} LoRA-MGPO & \textbf{86.58$_{\pm0.11}$} & \underline{94.72$_{\pm0.46}$} & \textbf{82.32$_{\pm0.18}$} & \textbf{93.79$_{\pm0.46}$} & \textbf{86.62$_{\pm0.68}$} & \textbf{88.81} \\
    \bottomrule
  \end{tabular}%
  \label{tab:addlabel}%
\end{table*}%

\section{Experiments}
\label{sec:experiments}

\subsection{Experimental Setup}

\paragraph{Baselines}
To provide a comprehensive evaluation, we compare LoRA-MGPO against a carefully selected set of baselines. These include Full Fine-Tuning (Full FT), serving as a strong performance benchmark, and vanilla LoRA \citep{hu2021lora}, our primary point of comparison. We further include two categories of state-of-the-art LoRA variants. The first category, \textit{variants with architectural modifications}, comprises methods that alter the LoRA structure itself, such as DoRA \citep{liu2024doraweightdecomposedlowrankadaptation}, which introduces learnable magnitude vectors, and AdaLoRA \citep{zhang2023adaloraadaptivebudgetallocation}, which dynamically allocates rank budgets. The second category, \textit{variants improving the training process or initialization}, includes rsLoRA \citep{kalajdzievski2023rankstabilizationscalingfactor}, which stabilizes update magnitudes; LoRA+ \citep{hayou2024loraefficientlowrank}, which employs different learning rates for the LoRA matrices; and PiSSA \citep{meng2024pissaprincipalsingularvalues}, which refines initialization using SVD. Finally, we compare against methods focused on \textit{gradient alignment}, such as LoRA-GA \citep{lora-ga} and LoRA-Pro \citep{wang2024loraprolowrankadaptersproperly}, which aim to align LoRA's gradient updates more closely with those of full fine-tuning.

\paragraph{Datasets}
Our experiments span a range of tasks in natural language understanding and generation. For NLU, we evaluate on five tasks from the widely-used General Language Understanding Evaluation (GLUE) benchmark~\citep{wang2018glue}: MNLI, SST-2, CoLA, QNLI, and MRPC. These tasks cover natural language inference, sentiment analysis, grammatical acceptability, and paraphrase identification.

For NLG, we fine-tune the LLaMA-2-7B~\citep{touvron2023llama} model on a 52k randomly sampled subset of the WizardLM dataset~\citep{xu2024wizardlm}. We evaluate the model on the MT-Bench dataset~\citep{zheng2024judging}, which consists of 80 multi-turn questions designed to assess conversational abilities across various aspects. The quality of the responses is evaluated by GPT-4, and we report the first-turn score as the primary evaluation metric.

For mathematical reasoning, we use a 100k random sample from MetaMathQA~\citep{yu2024metamath}, with evaluation on the GSM8K test set~\citep{cobbe2021training}. For code generation, fine-tuning is performed on a 100k randomly sampled subset of the CodeFeedback dataset~\citep{zheng2024opencodeinterpreter}, with evaluation on HumanEval~\citep{chen2021evaluating}.

\begin{table*}[htbp]
  \centering
\caption{Fine-tuning results of LLaMA-2-7B on MT-Bench, GSM8K, and HumanEval. Performance is evaluated using primary task metrics: MT-Bench score, GSM8K accuracy, and HumanEval Pass@1. PEFT methods are tested with rank $r=8$, and additional tests at ranks 32 and 128 are included to evaluate performance scaling. Results are averaged over three random seeds, with standard deviations provided. \textbf{Bold} and \underline{underlining} denote the best and second-best scores, respectively.}

    \begin{tabular}{l|ccc|c}
    \toprule
    \textbf{Method} & \textbf{MT-Bench} & \textbf{GSM8K} & \textbf{HumanEval} & \textbf{Avg} \\
    \midrule
    Full FT & 5.30$_{\pm0.11}$ & \textbf{59.36$_{\pm0.85}$} & \textbf{35.31$_{\pm2.13}$} & \textbf{33.32} \\
    LoRA  & 5.61$_{\pm0.10}$ & 42.08$_{\pm0.04}$ & 14.76$_{\pm0.17}$ & 20.82  \\
    
    \midrule
    
    DoRA  & \underline{5.97$_{\pm0.02}$} & 53.07$_{\pm0.75}$ & 19.75$_{\pm0.41}$ & 26.26  \\
    AdaLoRA & 5.57$_{\pm0.05}$ & 50.72$_{\pm1.39}$ & 17.80$_{\pm0.44}$ & 24.70  \\
    \midrule
    PiSSA & 5.30$_{\pm0.02}$ & 44.54$_{\pm0.27}$ & 16.02$_{\pm0.78}$ & 21.95  \\
    rsLoRA & 5.25$_{\pm0.03}$ & 45.62$_{\pm0.10}$ & 16.01$_{\pm0.79}$ & 22.29  \\
    LoRA+ & 5.71$_{\pm0.08}$ & 52.11$_{\pm0.62}$ & 18.17$_{\pm0.52}$ & 25.33  \\
    \midrule
    LoRA-GA & 5.95$_{\pm0.16}$ & 53.60$_{\pm0.30}$ & 19.81$_{\pm1.46}$ & 26.45  \\
    LoRA-GA (rank=32) & 5.79$_{\pm0.09}$ & 55.12$_{\pm0.30}$ & 20.18$_{\pm0.19}$ & 27.03  \\
    LoRA-GA (rank=128) & 6.13$_{\pm0.07}$ & 55.07$_{\pm0.18}$ & 23.05$_{\pm0.37}$ & 28.08  \\
    \midrule
   \rowcolor{green!20} LoRA-MGPO & \textbf{6.27$_{\pm0.12}$} & \underline{54.56$_{\pm0.44}$} & \underline{21.02$_{\pm0.39}$} & \underline{27.28} \\
   
    \rowcolor{green!20} LoRA-MGPO (rank=32) & 6.21$_{\pm0.15}$ & 55.74$_{\pm0.21}$ & 21.34$_{\pm0.47}$ & 27.76 \\

    \rowcolor{green!20} LoRA-MGPO (rank=128) & 6.48$_{\pm0.23}$ & 56.96$_{\pm0.35}$ & 24.87$_{\pm0.54}$ & 29.44\\
    \bottomrule
    \end{tabular}%
  \label{tab:addlabel2}%
\end{table*}%

\paragraph{Implementation Details}
For fair comparison, our experimental setup closely follows that of LoRA-GA~\citep{lora-ga}. Across all experiments, we use the AdamW optimizer~\citep{loshchilov2019decoupled} with weight decay set to 0 and a cosine learning rate schedule with a warm-up ratio of 0.03. LoRA adapters are applied to all linear layers within the transformer blocks, with the rank $r$ set to 8 and scaling factor $\alpha$ to 16 by default. For our two task families, the settings are as follows. For Natural Language Understanding (NLU) on GLUE, we fine-tune T5-base~\citep{raffel2020exploring} with a learning rate of $1 \times 10^{-4}$, a sequence length of 128, and a batch size of 32. The MGPO hyperparameters are $\rho=0.05$, $\mu=0.9$ (AdamW's `beta1`), and $\beta=0.9$. For Natural Language Generation (NLG), we fine-tune LLaMA-2-7B~\citep{touvron2023llama} with a learning rate of $2 \times 10^{-5}$ and a sequence length of 1024. We use a per-device batch size of 4 with 8 gradient accumulation steps for an effective batch size of 32. The MGPO hyperparameters are $\rho=0.01$, $\mu=0.8$ (AdamW's `beta1`), and $\beta=0.8$. All experiments were conducted on NVIDIA H20 96GB GPUs, repeated three times with different random seeds, and we report the average and standard deviation of the results. Further details on optimizer settings, specific LoRA target modules, and the software environment are provided in the Appendix.

\begin{figure*}[!h]
\centering
\includegraphics[width=.98\textwidth]{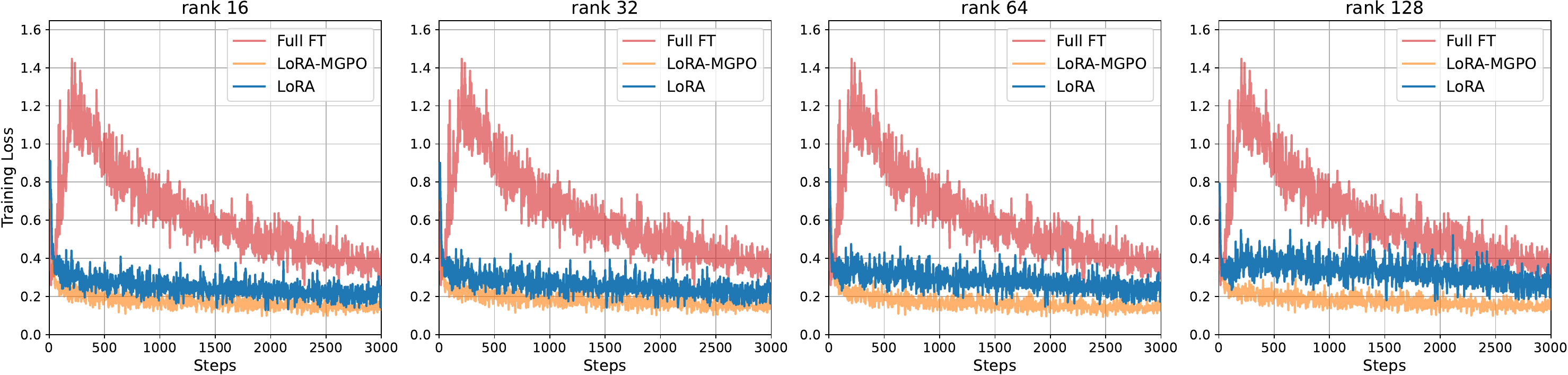}
\caption{Training loss dynamics across different rank configurations: A comparative analysis of LoRA, LoRA-MGPO, and full fine-tuning on LLaMA-2-7B with MetaMathQA. Rank ($r$) and alpha ($\alpha$) follow $r = \alpha \in \{16, 32, 64, 128\}$ with a fixed learning rate of $5e-4$.}
\label{fig: double}
\end{figure*}

\begin{figure*}[!h]
\centering
\includegraphics[width=.98\textwidth]{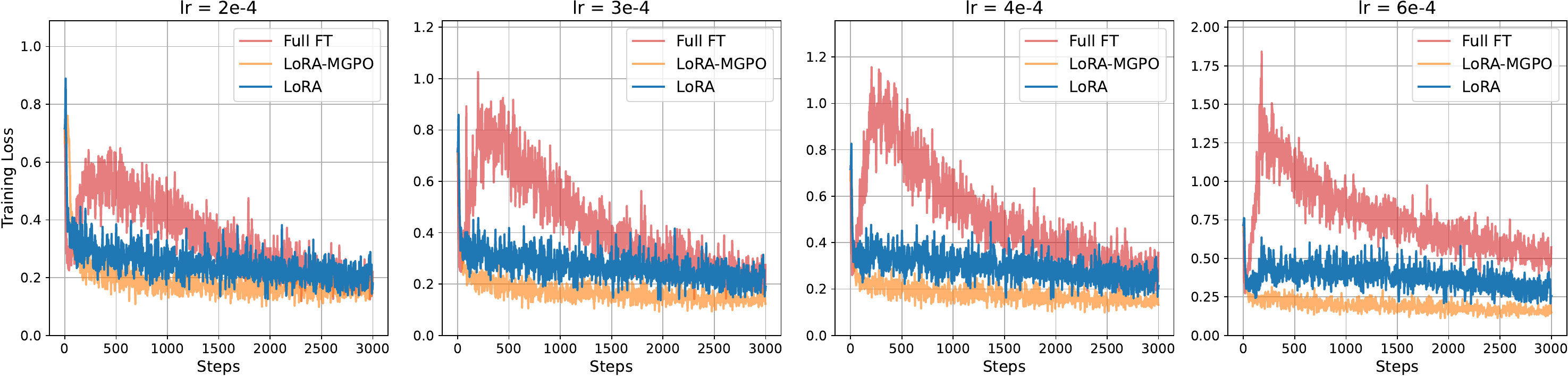}
\caption{Learning rate sensitivity analysis: A comparison of training loss for LoRA, LoRA-MGPO, and full fine-tuning on LLaMA-2-7B with MetaMathQA. The analysis spans learning rates $\{2e-4, 3e-4, 4e-4, 6e-4\}$, with rank ($r$) and alpha ($\alpha$) fixed at 128.}
\label{fig: lr_sensitivity}
\end{figure*}

\begin{figure*}[!h]
\centering
\includegraphics[width=.8\textwidth]{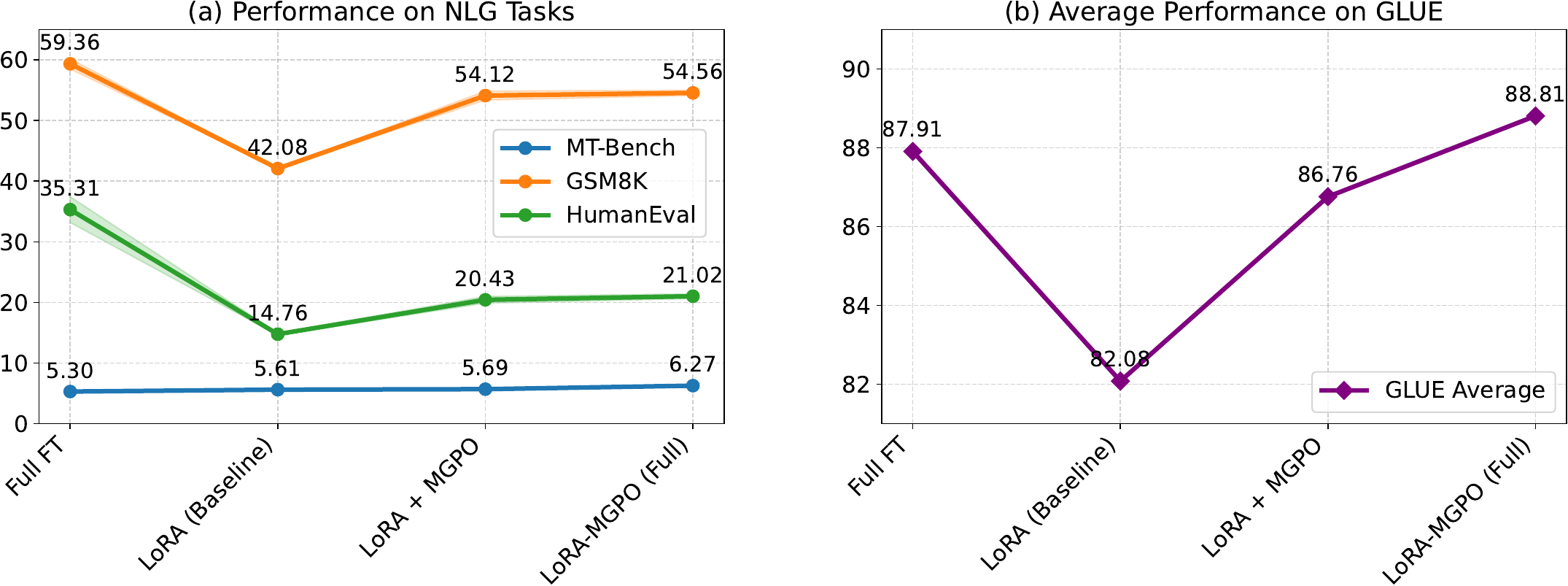}
\caption{Ablation study of LoRA-MGPO on NLG and NLU tasks. (a) LLaMA-2-7B performance across three NLG tasks. (b) T5-Base performance on the GLUE benchmark. "LoRA (Baseline)" refers to standard LoRA, "LoRA + MGPO" refers to an ablation with only momentum-guided perturbation, and "LoRA-MGPO (Full)" includes both momentum-guided perturbation and adaptive normalization.}
\label{fig:ablation}
\end{figure*}

\begin{table*}[htbp]
\centering
\caption{Comparison of computational efficiency and performance across LoRA, LoRA-MGPO, and Full FT methods, trained for one epoch on the WizardLM dataset using LLaMA-2-7B.}
\label{tab:method_comparison}
\resizebox{.96\linewidth}{!}{\begin{tabular}{l|c|c|c|ccl}
\toprule
Method & \#Params & Memory Cost & Training Time & MT-Bench& GSM8K&HumanEval\\
\midrule
Full FT & 6738M & >96 GB & - & $5.30_{\pm 0.11}$& $59.36_{\pm 0.85}$&$35.31_{\pm 2.13}$\\
LoRA & 320M & 81.73 GB & 5h 48min & $5.61_{\pm 0.10}$& $42.08_{\pm 0.04}$&$14.76_{\pm 0.17}$\\
 LoRA-MGPO & 320M & 90.56 GB & 6h 52min & $6.27_{\pm 0.12}$& $54.56_{\pm 0.44}$&$21.02_{\pm 0.39}$\\
\bottomrule
\end{tabular}}
\end{table*}
\begin{table}[h!]
\centering

\caption{Ablation study of LoRA-MGPO vs. random noise perturbation on three NLG benchmarks. Experiments use LLaMA-3.1-8B-Base \citep{dubey2024llama3} with rank $r=8$. Scores are averaged over three random seeds, with standard deviations in subscripts. \textbf{Bold} indicates the best method.}
\label{tab:ablation3}

\resizebox{\columnwidth}{!}{%
\begin{tabular}{l|c c c}
\toprule
\textbf{Method}                & \textbf{MTBench} & \textbf{GSM8k} & \textbf{HumanEval} \\ 
\midrule
Full FT                      & 5.88$_{\pm0.23}$  & 73.69$_{\pm0.28}$  & 51.63$_{\pm1.27}$ \\
LoRA                         & 6.15$_{\pm0.02}$  & 67.78$_{\pm1.25}$  & 43.09$_{\pm0.35}$ \\
LoRA + Random Noise              & 6.43$_{\pm0.26}$  & 68.05$_{\pm1.12}$  & 42.92$_{\pm0.41}$ \\
\textbf{LoRA-MGPO} & \textbf{7.51$_{\pm0.07}$}  & \textbf{70.23$_{\pm1.08}$}  & \textbf{45.13$_{\pm0.63}$} \\
\midrule
\end{tabular}%
}
\end{table}

\subsection{Main Results}

\paragraph{Performance on Natural Language Understanding (NLU)}
We first evaluated LoRA-MGPO on a standard suite of NLU tasks from the GLUE benchmark~\citep{wang2018glue}, using the T5-base model. As detailed in Table~\ref{tab:addlabel}, our method demonstrates strong and consistent performance. The improvements are particularly notable on challenging, low-resource benchmarks such as CoLA and MRPC, where LoRA-MGPO surpasses not only all other PEFT methods but also full fine-tuning. Success on these tasks often hinges on capturing subtle linguistic nuances. The stability afforded by LoRA-MGPO likely prevents the fine-tuning process from corrupting the rich knowledge encoded in the base model; by preventing erratic weight updates, our method may better preserve the pre-trained model's nuanced understanding of syntax and semantics. Quantitatively, LoRA-MGPO achieves the highest scores among all PEFT methods on five out of five tasks, obtains the best average score, and outperforms the next-best PEFT method, LoRA-GA, by a margin of 1.04 points.

\paragraph{Performance on Natural Language Generation (NLG)}
We further assessed our method on three challenging NLG tasks using the LLaMA-2-7B model, with results summarized in Table~\ref{tab:addlabel2}. LoRA-MGPO consistently secures top performance among all PEFT baselines. On the conversational MT-Bench, its top score suggests that stable training helps maintain the model's coherence and instruction-following capabilities. For structured reasoning tasks like mathematical problem-solving (GSM8K) and code generation (HumanEval), where logical consistency is paramount, LoRA-MGPO again emerges as the strongest PEFT method. A stable optimization trajectory may reduce the risk of the model deviating from a correct reasoning path during fine-tuning, as each update step is more measured, preventing catastrophic error accumulation common in multi-step generation. While full fine-tuning still holds an edge on the reasoning tasks, our method narrows the gap and outperforms it on MT-Bench. Notably, as the LoRA rank increases from 8 to 128, the performance of LoRA-MGPO scales gracefully, validating its ability to effectively leverage a higher parameter budget while maintaining the training stability that standard LoRA often lacks at higher ranks.

\subsection{Analysis and Ablation Studies}

\paragraph{Effectiveness in Mitigating Double Descent}
To empirically validate LoRA-MGPO's core claim of mitigating double descent, we conducted a controlled analysis of its training dynamics, focusing on the impacts of rank and learning rate. The results, presented in Figure~\ref{fig: double} and Figure~\ref{fig: lr_sensitivity}, offer compelling visual evidence of our method's stability. Figure~\ref{fig: double} illustrates that as the LoRA rank $r$ increases, the double descent phenomenon in standard LoRA becomes progressively more severe, exhibiting a sharp rebound at $r=128$. In stark contrast, LoRA-MGPO's loss curve remains smooth and monotonically decreasing across all ranks. Similarly, Figure~\ref{fig: lr_sensitivity} shows that while higher learning rates induce significant oscillations in standard LoRA, LoRA-MGPO maintains a stable convergence path. These findings provide strong empirical evidence that our method effectively stabilizes fine-tuning and potentially broadens the effective learning rate window.

\paragraph{Ablation Study}
To rigorously dissect the individual and combined contributions of our method's two key components—Momentum-Guided Perturbation (MGPO) and Adaptive Perturbation Normalization (APN)—we conducted a detailed ablation study, with results shown in Figure~\ref{fig:ablation}. The findings clearly validate our design choices. The first ablation step, labeled \textbf{`LoRA + MGPO`}, applies only the MGPO component and yields a substantial performance lift over the vanilla \textbf{`LoRA (Baseline)`}. On the NLU task suite, for instance, this single component boosts the average score from 82.08 to 86.76, demonstrating that the core strategy of using momentum to guide perturbations towards flatter loss regions is fundamentally effective.

However, the full potential is unlocked when introducing APN. Our complete model, labeled \textbf{`LoRA-MGPO (Full)`}, combines both components and achieves the final NLU score of 88.81. The significant improvement from 86.76 to 88.81 underscores the critical role of adaptive normalization. It suggests that while MGPO provides a stable perturbation \textit{direction}, its effectiveness is maximized only when the perturbation \textit{magnitude} is dynamically scaled in response to the gradient landscape. The consistent superiority of the full model across all NLU and NLG tasks confirms that these two components are not merely additive but work in synergy, fulfilling the design goals of our framework.

\paragraph{Comparison with Random Noise Perturbation}
To further validate that our performance gains stem from a principled optimization strategy rather than simple regularization, we compared LoRA-MGPO to LoRA augmented with undirected, isotropic random noise. The results in Table~\ref{tab:ablation3} are revealing: adding random noise provides only inconsistent and marginal benefits, and can even be detrimental in some cases (e.g., HumanEval). In contrast, LoRA-MGPO yields consistent and significant improvements across all tasks.

This disparity highlights a fundamental difference in mechanism. Random noise acts as a general regularizer by pushing parameters out of their immediate trajectory, which can occasionally help escape sharp minima by chance. However, the direction is arbitrary and uncorrelated with the loss landscape's structure. Our momentum-guided perturbation, conversely, is \textit{informed}. It leverages the recent history of the optimization path—a strong indicator of relevant high-curvature directions—to perform a targeted exploration. This principled approach makes the search for flat minima non-stochastic and significantly more effective and reliable than undirected noise injection.

\paragraph{Computational Cost Analysis}
Finally, we analyzed the practical overhead of our method (Table~\ref{tab:method_comparison}). As expected, LoRA-MGPO operates with the same minimal number of trainable parameters as standard LoRA, making it vastly more memory-efficient than Full FT. In terms of training time, LoRA-MGPO introduces a modest and acceptable overhead compared to vanilla LoRA (6h 52m vs. 5h 48m in our NLG setup). Given the significant performance improvements it delivers, this analysis confirms that LoRA-MGPO presents a highly favorable trade-off between computational cost and model performance, underscoring its practical viability.

\section{Related Work}

\paragraph{Parameter-Efficient Fine-Tuning (PEFT)}
The prohibitive computational and storage costs of full-parameter fine-tuning \citep{howard2018universal, devlin2018bert} have spurred the development of PEFT techniques for adapting large language models \citep{houlsby2019parameter, ding2023parameter}. By selectively updating a small subset of parameters, PEFT methods can achieve performance competitive with full fine-tuning while being significantly more efficient \citep{han2024parameter}. Among the diverse PEFT strategies, Low-Rank Adaptation (LoRA) \citep{hu2021lora} has gained prominence for its simplicity and effectiveness. Recent works have enhanced LoRA along several directions. One line of work introduces \textit{architectural modifications}; for instance, DoRA \citep{liu2024doraweightdecomposedlowrankadaptation} integrates learnable magnitude vectors, while AdaLoRA \citep{zhang2023adaloraadaptivebudgetallocation} dynamically allocates rank budgets. Another direction focuses on \textit{improving the training process and initialization}, such as adjusting scaling factors in rsLoRA \citep{kalajdzievski2023rankstabilizationscalingfactor}, using separate learning rates in LoRA+ \citep{hayou2024loraefficientlowrank}, or refining initialization with PiSSA \citep{meng2024pissaprincipalsingularvalues} and NLoRA \citep{guo2025nlora}. A third direction aims to improve the quality of the parameter updates, for instance by alleviating training biases with BA-LoRA \citep{chang2024ba} or by more closely aligning LoRA's gradients with those of full fine-tuning, as seen in LoRA-GA \citep{lora-ga} and LoRA-Pro \citep{wang2024loraprolowrankadaptersproperly}. Additional work has further explored LoRA's application in multi-task learning, such as \citep{liu2025align,liu2025r}. Distinct from these approaches, our work focuses directly on the underlying optimization dynamics. Rather than altering LoRA's architecture or mimicking full fine-tuning gradients, we introduce a novel training framework to stabilize the optimization process itself.

\paragraph{Optimization Stability in PEFT}
The training stability of PEFT methods, particularly LoRA, is a critical concern. Empirical studies have revealed that as LoRA's rank increases, performance can degrade after an initial improvement, a behavior analogous to the double descent phenomenon \citep{belkin2019reconciling, nakkiran2019deepdoubledescentbigger}. This instability highlights the challenge of navigating high-dimensional and non-convex loss landscapes during fine-tuning. To promote smoother optimization and find flatter minima, Sharpness-Aware Minimization (SAM) \citep{foret2020sharpness} has been influential. However, its requirement for dual gradient computations imposes a significant computational burden \citep{becker2024momentum, li2024revisiting}. More recent work has explored more efficient directional perturbation strategies. Momentum-guided methods, for example, reuse optimizer momentum to avoid the extra gradient step, reducing computational cost without sacrificing the directional guidance \citep{becker2024momentum}. Other techniques, such as applying an exponential moving average (EMA) to model weights, also contribute to stability by smoothing the trajectory of parameter updates \citep{wang2021rethinking}. While these components---efficient perturbation and smoothing---are individually effective, they are typically studied in isolation. This leaves a clear gap for a unified framework that synergistically combines these strategies to enhance both the efficiency and stability of PEFT. Our work, LoRA-MGPO, is designed to fill this gap.

%%%%%%%%%%%%%%%%%%%%%%%%%%%%%%%%%%%%%%

\section{Conclusion}
\label{sec:conclusion}

In this work, we addressed the double descent phenomenon in Low-Rank Adaptation (LoRA), an instability that can affect the fine-tuning of large language models. We proposed \textbf{LoRA-MGPO}, an optimization framework that integrates Momentum-Guided Perturbation Optimization (MGPO). This method aims to find flatter minima by reusing optimizer momentum to guide weight perturbations, combined with an adaptive normalization scheme to improve robustness. Our experimental results across a range of natural language understanding (NLU) and natural language generation (NLG) tasks show that LoRA-MGPO provides improved performance over standard LoRA and other common PEFT baselines. This improvement is reflected in more stable convergence trajectories and reduced training instability. LoRA-MGPO offers a practical approach to overcoming some of the optimization challenges in LoRA while maintaining its parameter efficiency. Future research may explore extending this framework to other parameter-efficient methods or adapting it for different domains, such as vision and speech.

\section*{Limitations}

First, LoRA-MGPO's use of momentum vectors for perturbation directions assumes relatively stable optimizer dynamics, which might limit its effectiveness during early training stages or in the presence of highly non-stationary gradient conditions. Second, while the adaptive perturbation normalization via EMA-smoothed gradients improves robustness, its performance may be sensitive to sudden changes in gradient magnitude distributions, potentially requiring adjustments to the smoothing hyperparameters depending on the specific task.

\section*{Ethics Statement}
Our research focuses on LoRA-MGPO, a general-purpose optimization algorithm designed to improve the stability of parameter-efficient fine-tuning (PEFT). The experiments use publicly available, pre-trained models (LLaMA-2-7B, T5-base) and standard academic benchmarks. We acknowledge that these foundational models may inherit and potentially amplify societal biases present in their training data. The primary goal of this work is to provide a more reliable and resource-efficient tool for adapting and studying such models within the research community. By enhancing PEFT techniques, our work contributes to broader efforts aimed at reducing the computational costs involved in large-scale model adaptation.

\section*{Acknowledgments}
This work is supported by the National Key Research and Development Program of China (No.2023YFF0905400), the National Natural Science Foundation of China (No.U2341229) and the Reform Commission Foundation of Jilin Province (No.2024C003).

\bibliography{refs}

\newpage
\appendix

\vspace{2em}
\begin{center}
    \Large{\textbf{Appendix}}
\end{center}
\vspace{2em}

\etocdepthtag.toc{appendix}
\etocsettagdepth{chapter}{none}
\etocsettagdepth{appendix}{subsection}
\tableofcontents

\section{Models and Datasets}

\subsection{Details of Models}
In this work, we primarily utilize two pre-trained language models: LLaMA-2-7B and T5-base. 

\begin{itemize}
    \item \textbf{LLaMA-2-7B}: A 7-billion parameter, decoder-only transformer model from the LLaMA-2 series, primarily used for generation tasks. More details are available at its Hugging Face repository\footnote{\url{https://huggingface.co/meta-llama/LLaMA-2-7B}}.
    
    \item \textbf{T5-base}: A 220-million parameter encoder-decoder transformer model, widely used for a variety of natural language understanding tasks. More details are available at its Hugging Face repository\footnote{\url{https://huggingface.co/t5-base}}.
\end{itemize}

Our experiments were conducted using the implementations of these models provided by the Hugging Face Transformers library.

\begin{table*}[!ht]
\centering
\caption{GLUE Benchmark Datasets and Evaluation Metrics}
\label{tab:glue_datasets}
\resizebox{.98\textwidth}{!}{%
\begin{tabular}{llcccc}
\toprule
\textbf{Dataset} & \textbf{Task Type} & \textbf{Classes} & \textbf{Train Examples}& \textbf{Metric} & \textbf{Description} \\
\midrule
CoLA   & Acceptability   & 2       & 8.5k    & Matthews Corr.          & Grammatical acceptability \\
SST-2  & Sentiment       & 2       & 67k     & Accuracy                & Sentiment analysis \\
MRPC   & Paraphrase      & 2       & 3.7k    & Accuracy/F1             & Paraphrase detection \\
MNLI   & NLI             & 3       & 393k    & Accuracy                & Multi-genre NLI \\
QNLI   & NLI/QA          & 2       & 108k    & Accuracy                & QA/NLI converted from SQuAD \\
\bottomrule
\end{tabular}%
}
\end{table*}

\subsection{Details of Datasets}
Table~\ref{tab:glue_datasets} summarizes the GLUE benchmark datasets \citep{wang2018glue}. For our Natural Language Generation (NLG) experiments, we used the following evaluation metrics: Accuracy for GSM8K; Pass@1 for HumanEval; and a score based on GPT-4 evaluation for MT-Bench.

\section{Baselines and Implementation}

\subsection{Baseline Methods}
Our study includes several baseline methods for a comprehensive comparison. \textbf{Full Fine-Tuning} serves as a strong performance benchmark. \textbf{Vanilla LoRA} \citep{hu2021lora} is our primary point of comparison from the PEFT literature. We also compare against LoRA variants that introduce \textbf{structural modifications} (DoRA \citep{liu2024doraweightdecomposedlowrankadaptation}, AdaLoRA \citep{zhang2023adaloraadaptivebudgetallocation}) and those that \textbf{refine the training process or initialization} (rsLoRA \citep{kalajdzievski2023rankstabilizationscalingfactor}, LoRA+ \citep{hayou2024loraefficientlowrank}, PiSSA \citep{meng2024pissaprincipalsingularvalues}). Finally, we include methods focused on \textbf{gradient alignment} (LoRA-GA \citep{lora-ga}, LoRA-Pro \citep{wang2024loraprolowrankadaptersproperly}).

\subsection{Implementation Details}
\paragraph{LoRA Configuration.} As stated in the main text, LoRA adapters were applied to all linear layers within the transformer blocks for both LLaMA-2-7B and T5-base models.

\paragraph{Initialization of MGPO.} The implementation of our method requires an initial state for the momentum vector and the adaptive normalization factor. Following standard optimizer practice, the momentum `$\boldsymbol{m}$` is initialized to zeros. The adaptive normalization factor `$\bar{g}$` is initialized using the L2-norm of the gradient computed in the first training step.

\paragraph{Hyperparameters.} Our method introduces two primary hyperparameters: the perturbation radius `$\rho$` and the EMA decay rate `$\beta$`. `$\rho$` controls the magnitude of the weight perturbation, influencing the search for flatter minima. `$\beta$` controls the temporal smoothing window for the adaptive normalization. The values used in our main experiments were effective across the evaluated tasks, as evidenced by the strong performance reported in Section~\ref{sec:experiments}.

\subsection{Hyperparameter Settings for Baselines}
To ensure a fair and robust comparison, we adhered to the hyperparameter settings recommended in the original papers or official codebases of our baseline methods wherever possible. General settings, such as the learning rate schedule and batch size, were kept consistent across all methods as described in Section~\ref{sec:experiments}. Key method-specific hyperparameters are detailed below.

\begin{itemize}
    \item \textbf{DoRA \citep{liu2024doraweightdecomposedlowrankadaptation}:} We utilized the official implementation provided by the authors, maintaining its default configuration for the magnitude and directional components.
    
    \item \textbf{AdaLoRA \citep{zhang2023adaloraadaptivebudgetallocation}:} We followed the setup from the original paper, with the rank budget dynamically allocated starting from a higher initial rank and pruned during training.
    
    \item \textbf{LoRA+ \citep{hayou2024loraefficientlowrank}:} Following the authors' recommendation, the learning rate for the LoRA matrix $A$ was set to our default value ($1 \times 10^{-4}$ for NLU, $2 \times 10^{-5}$ for NLG), while the learning rate for matrix $B$ was set 16 times higher.
    
    \item \textbf{LoRA-GA and LoRA-Pro \citep{lora-ga, wang2024loraprolowrankadaptersproperly}:} For these methods focused on gradient alignment, we used the hyperparameter settings as specified in their respective papers and official implementations to ensure a faithful comparison.
\end{itemize}

For all other baselines, we used their standard, publicly available implementations without modification to their core components.

\end{document}